\documentclass[letterpaper, 10 pt, journal, twoside]{IEEEtran}

\IEEEoverridecommandlockouts                              % This command is only needed if 
\usepackage{amsmath}
\usepackage{amssymb}
\usepackage{tabularx}

\usepackage[final]{pdfpages}
\usepackage{algorithm}
\usepackage{algpseudocode}
\usepackage{booktabs}
\usepackage[short]{optidef}
\usepackage{soul,xcolor}
\usepackage{siunitx}
\usepackage{import}
\usepackage{amsthm}
\usepackage{bm}
\usepackage{tablefootnote}
\usepackage{physics}

\usepackage{units}
\usepackage{arydshln}
\usepackage{lipsum}

\usepackage{url}
\usepackage{hyperref}
\usepackage{balance}

\usepackage{graphicx}

\newlength\figureheight
\newlength\figurewidth

\begin{document}
%==============================================================

%==============================================================
%TITLE PAGE
%==============================================================

\title{Optimization-Based Hierarchical Motion Planning for Autonomous Racing}
\author{Jos\'e~L.~V\'azquez$^{1,*}$, Marius~Br\"uhlmeier$^{1,*}$, Alexander~Liniger$^{2,*}$, Alisa~Rupenyan$^{1}$, John~Lygeros$^{1}$
\thanks{$^{1}$ Automatic Control Lab, ETH Zurich, Switzerland \newline
 	{\tt{\footnotesize \{vjose,mariusbr\}@student.ethz.ch, \newline \{ralisa,lygeros\}@control.ee.ethz.ch}}}%
\thanks{$^{2}$ Computer Vision Lab, ETH Zurich, Switzerland \newline {\tt \footnotesize alex.liniger@vision.ee.ethz.ch}}
\thanks{$^*$ The authors contributed equally.}
}
\maketitle

%==============================================================

\begin{abstract}
    In this paper we propose a hierarchical controller for autonomous racing where the same vehicle model is used in a two level optimization framework for motion planning. The high-level controller computes a trajectory that minimizes the lap time, and the low-level nonlinear model predictive path following controller tracks the computed trajectory online. Following a computed optimal trajectory avoids online planning and enables fast computational times. The efficiency is further enhanced by the coupling of the two levels through a terminal constraint, computed in the high-level controller. Including this constraint in the real-time optimization level ensures that the prediction horizon can be shortened, while safety is guaranteed. This proves crucial for the experimental validation of the approach on a full size driverless race car. The vehicle in question won two international student racing competitions using the proposed framework; moreover, our hierarchical controller achieved an improvement of 20\% in the lap time compared to the state of the art result achieved using a very similar car and track.
\end{abstract}

\section{Introduction}
\begin{figure}[t]
\includegraphics[width=0.45\textwidth]{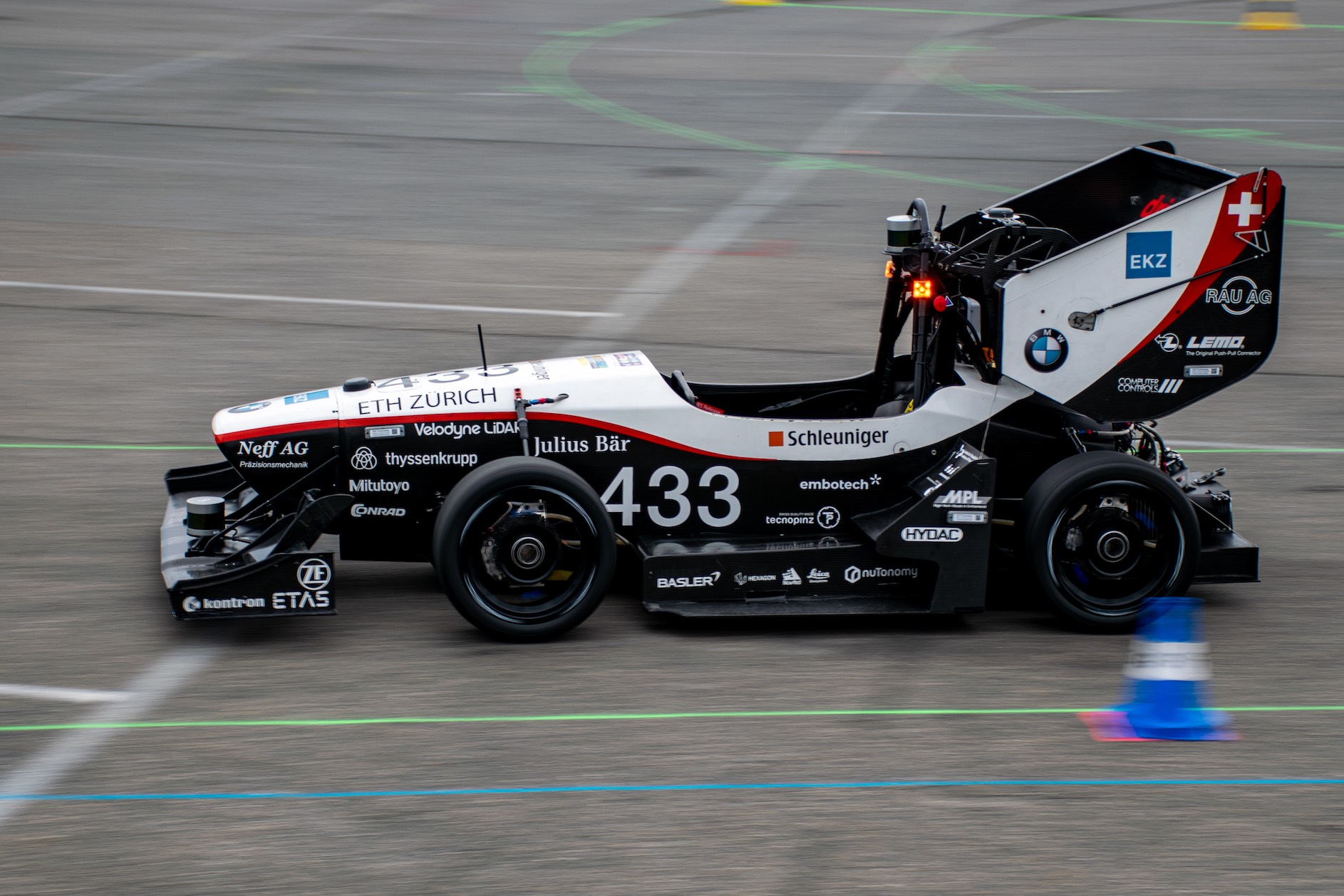}
\centering
\caption{Photo of \textit{pilatus} \textcopyright FSG Gosala}
\label{fig:pilatus}
\end{figure}

Over the past few decades motion planning for autonomous driving has attracted great interest in both academic and industrial research, see for example \cite{paden2016survey} for a review on motion planning, and \cite{buehler2005,buehler2009} for a robotic system perspective. A sub-field of autonomous driving where motion planning is of crucial importance, is autonomous racing, where the goal is to drive around a given race track as fast as possible. This requires a motion planning method that pushes the car to the limit of handling, and is able to handle the nonlinear behavior in this regime \cite{Gerdes2012,Liniger2015}. In this paper, we propose a novel hierarchical optimization-based motion planner that exploits the structure of the given task, and demonstrate the performance using the Formula Student Driverless (FSD) car developed at ETH Zurich. The robotic platform with the name \textit{pilatus} is shown in Fig. \ref{fig:pilatus}, and won two international autonomous racing competitions using the algorithms presented in this work.

When investigating existing autonomous racing motion planners and controllers, three classes can be distinguished: The first class computes an ideal path and a velocity profile around the track offline, and then follows this path using a static feedback controller \cite{Gerdes2012,Betz2019}. These approaches are intriguing due to their simplicity but since all motion planning is done offline they lack flexibility. The remaining two classes rely on online optimization-based motion planning \cite{Liniger2015,funke2016collision}. The methods are interesting because they are able to consider the nonlinear vehicle dynamics, track constraints, input constraints, and at the same time plan an optimal motion for the vehicle. The challenge however is that autonomous racing requires fast sampling times in the order of 10-50ms, while finding the optimal trajectory around a race track requires significant foresight in the order of several seconds. This can be challenging since the model complexity and planning horizon both contribute to increase the computational burden. The main difference between the two online optimization based classes lies exactly in addressing this problem. One is based purely on online motion planning. The advantage is that it runs completely online, however, given today's computational platforms, these methods are limited to special applications such as miniature race cars \cite{Liniger2015,verschueren}. Extension to full size car is possible \cite{Kabzan2019_AMZ}, but only for limited top speeds. The last approach strikes a balance between the two previous classes and combines both offline and online motion planning \cite{LinigerTCST,Novi2019,caporale2019towards,Rosolia2017}. Our work falls under this third class that mixes online and offline elements; the building blocks of the method are outlined below.

In \cite{LinigerTCST} a viability constraint is computed offline that guides the online motion planner. In \cite{Rosolia2017} terminal constraints and cost are estimated using experimental data, these terminal ingredients then guide the online Nonlinear Model Predictive Control (NMPC) motion planner. In \cite{caporale2019towards}, a least curvature path as well as velocity profile is computed offline, this trajectory is then tracked with simple NMPC. Finally, \cite{Novi2019} computes a least curvature path offline. To follow the path the method first computes a velocity profile that is then used as a terminal velocity constraint in a low-level NMPC motion planner based on a sophisticated vehicle model. Thus, \cite{Novi2019} combines the ideas of terminal constraints with an offline computed ideal line.

Our approach is mostly inspired by \cite{Novi2019}, specifically, our hierarchical motion planner computes a time-optimal trajectory around the race track offline, and a low-level NMPC follows this path, while restricted by a terminal velocity constraint extracted from the time-optimal trajectory. However, compared to \cite{Novi2019} our high-level motion planner computes a full time optimal state-input trajectory using a lap time optimization approach similar to \cite{lot2014curvilinear,Rucco2015}. This makes it possible to use the same model for the two levels, which guarantees that the minimum-time trajectory is drivable while achieving lower lap times. Compared to \cite{Novi2019} the model used in the high-level planner is more complex, while that in the lower level is simpler. This resulted in reducing significantly the online computational load, which enables the real-time implementation of our method. 

Our work makes the following contributions to the state of the art: First we propose a hierarchical controller where the two levels use the same vehicle model formulated in curvilinear coordinates. Second, we show that for the given FSD setting, following the minimum time trajectory improves performance compared to planning the motion from scratch, and that using terminal velocity constraints allows the use of shorter prediction horizons without sacrificing performance. Finally, we show that the proposed hierarchical controller can race a full-size autonomous FSD race car, finishing fastest in all the attended FSD competitions, and reducing the lap times shown in \cite{Kabzan2019_AMZ} by 20\%, using a similar car and track. See \url{https://youtu.be/gcnngFyWnFQ?t=13079} for the run at the formula student Germany competition and \url{https://youtu.be/vkVBi9LWJo0} for additional experiments and visualizations of the proposed method.

In Section \ref{sec:model} we introduce the vehicle model and constraints used in our motion planner, and discuss time and space-dependent curvilinear models. In Section \ref{sec:control} we formulate the high-level lap time optimization problem and the low-level NMPC path following problem. We show numerical and experimental results in Section \ref{sec:results}, and conclude the paper and give future direction of research in Section \ref{sec:conclusion}.

\section{Model}
\label{sec:model}
We first introduce the vehicle model used in the hierarchical curvilinear motion planner, including the model-related constraints, and the coordinate system. This allows us to highlight similarities and differences in the two layers of our control structure. First the model formulated in curvilinear coordinates is presented, followed by the model-related constraints, and finally the difference between time and space-dependent models is discussed. 

The starting point is a bicycle model with dynamic force laws \cite{jazar2008vehicle}. The main advantage of this model is that it strikes a balance between simplicity and accuracy. It is significantly simpler than a full-fledged double-track model with complex tire forces \cite{Novi2019}, while still modeling important effects such as tire slip and saturation.  

\subsection{Curvilinear Bicycle Model}
\begin{figure}[t]
    \centering
    \includegraphics[width=0.4\textwidth]{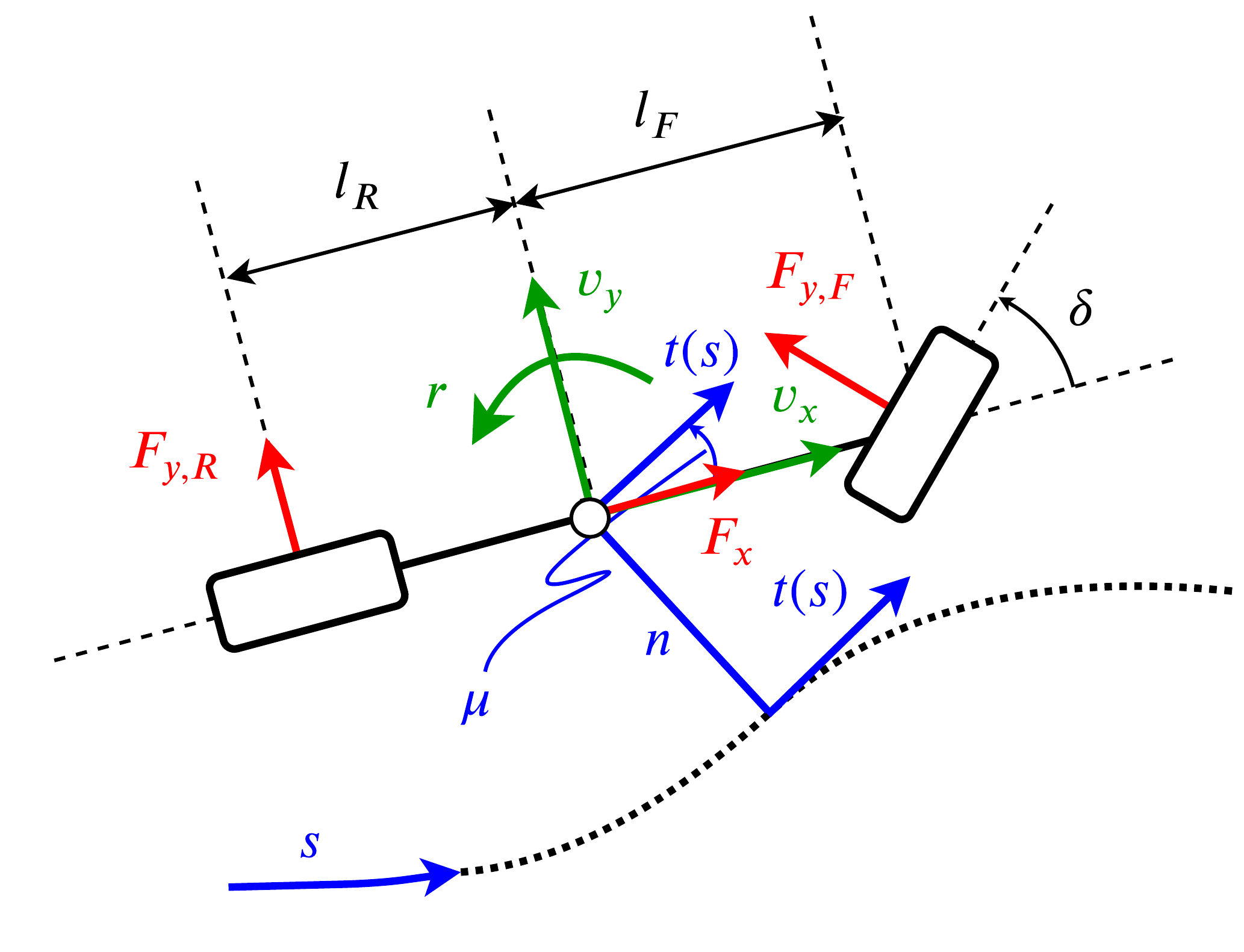}
    \caption{Illustration of the bicycle model in the curvilinear frame. The figure shows the curvilinear coordinates in blue, the velocity states in green and the forces red. The tangent of the reference path is presented as $t(s)$.}
    \label{fig:model}
\end{figure}

The bicycle model dynamics is set in curvilinear coordinates, which are formulated locally with respect to a given reference path. In our case the reference path can be the center line of the race track, or an ideal line. By using curvilinear coordinates, no global position or heading is considered, and the state information is given relative to the reference path. The state comprises of the progress (arc-length) along the path $s$, the orthogonal deviation from the path $n$, and the local heading $\mu$. Given the reference path the global coordinates can be recovered, and the curvilinear coordinates can be obtained (locally) by projecting the global coordinates onto the reference path. The bicycle model and the curvilinear coordinates are shown in Fig. \ref{fig:model}. 

Several assumptions are needed to formulate our vehicle model: (i) the car travels on level ground, (ii) load changes can be neglected, (iii) combined slip can be neglected, (iv) all the drive train forces act at the center of gravity, and (v) the vehicle is a rigid body with mass $m$ and rotational inertia $I_z$. The first assumption is valid as formula student race tracks are level. The remaining assumptions hold since we use a low level traction controller, that distributes the requested drive train force and torque vectoring moment to the four wheels while considering combined slip and load changes, similar to \cite{Kabzan2019_AMZ}.

The dimensions of the vehicle are given by $l_F$ and $l_R$ which describe the distance between the CoG and the corresponding front and rear axles. To formulate the dynamics we first consider the curvilinear states $s$, $n$, and $\mu$, and use the standard approach to integrate the velocities, which relies on the curvature $\kappa(s)$ of the reference path, see \cite{Rucco2015} for a detailed discussion. We consider next the longitudinal and lateral velocity $v_x$ and $v_y$ and the yaw rate $r$, which are influenced by the tire and drive train forces. Finally, we also consider the actuator dynamics by using an integrator model for the steering angle $\delta$ and the driver command $T$. Thus, the state is given by $\mathbf{\tilde{x}} = [s; n; \mu; v_x; v_y; r; \delta; T]$ and the input as $\mathbf{u} = [\Delta \delta, \Delta T]$. Under the assumptions listed above the dynamics then becomes

\begin{align}
    \dot{s} &= \dfrac{v_x\cos{\mu}-v_y\sin{\mu}}{1-n \kappa(s)}\,, \nonumber\\
    \dot{n} &= v_x\sin{\mu}+v_y\cos{\mu}\,, \nonumber\\
    \dot{\mu} &= r - \kappa(s) \dot{s}\,, \nonumber \\
    \dot{v}_x &= \tfrac{1}{m}(F_x - F_{y,F}\sin{\delta} + m v_y r)\,, \nonumber \\
    \dot{v}_y &= \tfrac{1}{m}(F_{y,R} + F_{y,F}\cos{\delta} - m v_x r) \,, \nonumber\\
    \dot{r} &= \tfrac{1}{I_z}(F_{y,F}l_F\cos{\delta} - F_{y,R}l_R + M_{tv})\,, \nonumber \\
    \dot{\delta} &= \Delta \delta \,, \nonumber\\
    \dot{T} &= \Delta T\,. \label{eq:model}
\end{align}

Here $\kappa(s)$ is the curvature at the progress $s$, $F_{y,F/R}$ are the lateral tire forces at the front and rear wheel and $F_{x}$ is the longitudinal force acting on the car. Finally, $M_{tv} = p_{tv}(r_\text{t} - r)$ is the moment the torque vectoring system produces, where $p_{tv}$ is the gain of the system and $r_\text{t} = \tan(\delta)v_x/(l_R + l_F)$ is the yaw rate target, see \cite{Kabzan2019_AMZ} for more details. Note that the curvature contains the full information about the reference path. To simplify the notation we denote the dynamics in \eqref{eq:model} by $\mathbf{\dot{\tilde{x}}} = f^c_t(\mathbf{\tilde{x}},\mathbf{u})$, where the superscript $c$ highlights that it is a continuous time model and the subscript $t$ that it is a time-domain model.

The tire forces are the most important part of the model, since they model the interaction of the car with the ground. The lateral tire forces $F_{y,F}$ and $F_{y,R}$ describe the lateral forces acting on the front, and respectively the rear tires. They are computed using a simplified Pacejka tire model \cite{pacejka1992magic},
\begin{align*}
    \alpha_F & = \arctan{\left( \frac{v_y + l_F r}{v_x} \right)} - \delta \,,\\
    \alpha_R & = \arctan{\left( \frac{v_y - l_R r}{v_x} \right)} \,,\\
    F_{y,F} & = F_{N,F} D_F\sin{(C_F\arctan{(B_F\alpha_F)})} \,,\\
    F_{y,R} & = F_{N,R} D_R\sin{(C_R\arctan{(B_R\alpha_R)})} \,,
\end{align*}
where $\alpha_{F/R}$ are the slip angles at the front and, respectively rear wheel, and $B_{F/R}$, $C_{F/R}$ and $D_{F/R}$ are the parameters of the simplified Pacejka tire model. Finally, the normal loads on the tires are given by $F_{N,F} = l_R/(l_F + l_R) m g$ and $F_{N,F} = l_F/(l_F + l_R) m g$.

The longitudinal drive train force $F_x$ is composed of a motor force $F_M = C_m T$, which is a linear function of the driver command $T \in [-1,1]$, as well as a constant rolling resistance term $C_{r0}$ and drag-force term $C_{r2} v_x^2$, thus we have $F_x = C_m T - C_{r0} - C_{r2}v_x^2\,$.
\subsection{Constraints}
\label{sec:constraints}
During the race, it is crucial to ensure that the entirety of the vehicle stays within the track boundaries. This is enforced via heading-dependent constraints on the lateral deviation $n$,
\begin{align}
\begin{split}
    n-\frac{L_c}{2}\sin{\abs{\mu}} +\frac{W_c}{2}cos{\mu} & \leq \mathcal{N}_L(s)\,,\\
    -n+\frac{L_c}{2}\sin{\abs{\mu}} + \frac{W_c}{2}\cos{\mu} & \leq \mathcal{N}_R(s)\,,
\end{split}
\label{eq:boundary}
\end{align}
where $L_c$ is the length and $W_c$ is the width of the car, and left and right track width at a specific progress $s$ are given by $\mathcal{N}_{R/L}(s)$. We denote the resulting track constraints \eqref{eq:boundary} as $\mathbf{\tilde{x}} \in \mathcal{X}_{\text{Track}}$.

Since the used tire model neglects combined tire forces, we use a friction ellipse constraint to limit the combined tire forces, such that the low level traction controller can better deal with the requested forces. The resulting friction ellipse constraints are given by
\begin{align}
    \left(\rho_{long}F_{M,F/R}\right)^2 + F_{y,F/R}^2 & \leq (\lambda D_{F/R})^2\,,
\label{eq:ellipse}
\end{align}
where $F_{M,F} = F_{M,R} = F_M /2$ is the motor force at the front, and respectively at the rear wheel, and a static 50/50 force split is assumed between the wheels. $\rho_{long}$ determines the shape of the ellipse, and controls how much force can be applied in the longitudinal direction, and $\lambda$ allows to determine the maximum combined force. We denote the friction ellipse constraints in \eqref{eq:ellipse} by $\mathbf{\tilde{x}} \in \mathcal{X}_\text{FE}$.

Finally, we also impose constraints on the physical inputs $\delta$ and $T$, as well as on the rate of change of these inputs $\Delta \delta$ and $\Delta T$. These constraints are simple box constraints and are determined based on physical limits for the steering input and safety considerations for the driving command. For a slim definition we define $\mathbf{a} = [\delta;T;\Delta \delta; \Delta T]$, and the box constraints are denoted as $\mathbf{a} \in \mathcal{A}$.

\subsection{Space vs Time Domain}
\label{sec:spacevstime}

A natural way to formulate dynamical systems is to describe them as a function of time, especially for the purpose of control where control inputs are applied with a fixed sampling frequency. However, for our dynamics it is also possible to formulate them to evolve with respect to space, making the state a function of the progress $\mathbf{\tilde{x}}(s)$. This transformation can be performed using the reformulation, 
\begin{align}
    &\mathbf{\dot{\tilde{x}}} = \frac{\partial \mathbf{\tilde{x}}}{\partial t} = \frac{\partial \mathbf{\tilde{x}}}{\partial s}\frac{\partial s}{\partial t} \nonumber\\
    & \Rightarrow \frac{\partial \mathbf{\tilde{x}}}{\partial s} = \frac{1}{\dot{s}}f^c_t(\mathbf{\tilde{x}}(s),\mathbf{u}(s))  = f^c_s(\mathbf{\tilde{x}}(s),\mathbf{u}(s)) \label{eq:spatialdomain}
\end{align}

Expressing the state as a function of $s$ comes with several advantages, for example the time is now a function of the states $t = 1/\dot{s}$. This allows us to naturally formulate a minimal time problem, as the time it takes to move from $s = s_0$ to $s_T$ is given by $T = \int_{s_0}^{s_T} 1/\dot{s} \; d\sigma$. One additional advantage is that the curvature does not change based on the inputs, since $s$ is no longer a function of time. This is especially relevant for long prediction horizons where time dependent formulations need an accurate initial guess of the solution. The last advantage is that the progress state $s$ becomes redundant and can be dropped from the state vector, thus a reduced state vector can be used for the space-curvilinear model  $\mathbf{{x}} = [n;\mu;v_x;v_y;r;\delta;T]$. The associated disadvantages are that first, when discretizing the system it is not time-discrete but space-discrete. This implies that, when using this model for control, inputs should not be changed with a fixed sampling time but at a fixed space interval. Second, pre-multiplying the system with $1/\dot{s}$ makes the dynamics computationally more complex for an optimization solver, since computing Jacobians becomes more expensive. Finally, the space-domain transformation introduces a singularity if the car stops ($\dot{s} = 0$). 

\section{Hierarchical Controller}
\label{sec:control}

\begin{figure}[h]
\includegraphics[width=0.45\textwidth]{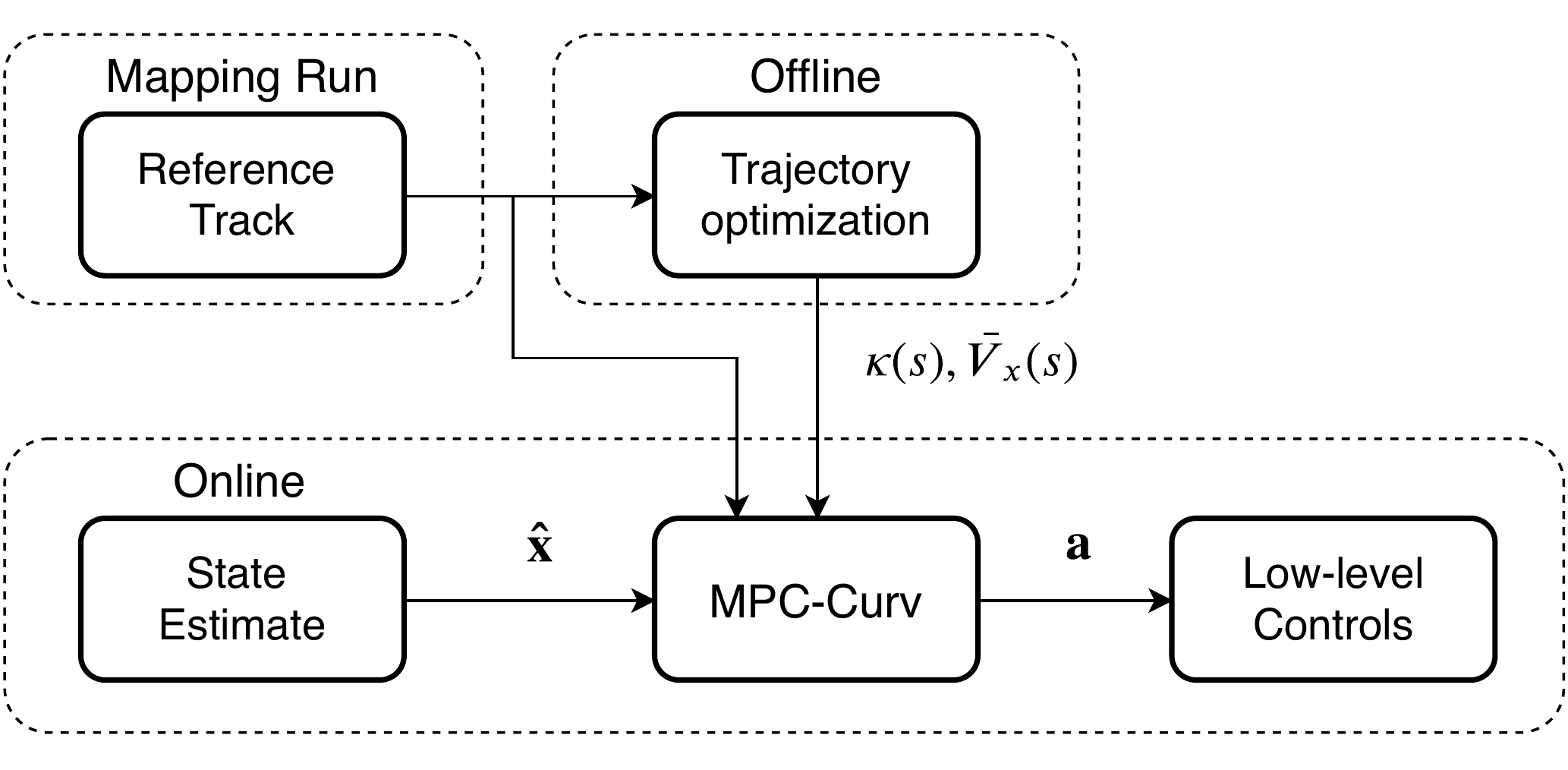}
\centering
\caption{The hierarchical controller uses the reference track in both stages of the hierarchical controller. First the lap time optimization (LTO) problem computes a reference path, whose curvature $\kappa(s)$ and speed profile $\bar{V}_x(s)$ are later used by the NMPC.}
\label{fig:diagram}
\end{figure}

One fundamental challenge when implementing an NMPC is the computational burden caused by solving an optimization problem at each time step. The main factors that influence the computation time are the number of states and inputs, and the prediction horizon. By using curvilinear coordinates, the number of states can be reduced compared to the current state of the art method \cite{Kabzan2019_AMZ}. To reduce the horizon length we first compute the Lap Time Optimization (LTO) trajectory, and then follow this path with a NMPC where the longitudinal speed at the end of the horizon is upper bounded by the LTO speed, see Fig. \ref{fig:diagram}. Note that the LTO path as well as the LTO terminal velocity are fundamental for shortening the horizon.  

An alternative view on this approach is in terms of terminal constraint and recursive feasibility. Since the same model is used for the two levels, and the LTO trajectory is periodic, the LTO trajectory is an invariant set of the system dynamics and the constraints. Thus, if the NMPC exactly steers the system to the LTO trajectory there exists an input sequence that keeps the system within the constraint set indefinitely. The terminal velocity constraint used here is a relaxation, compared to forcing the NMPC to exactly steering the system to the LTO trajectory. Thus, our approach does not guarantee recursive feasibility but comes with some advantages. For one, we avoid a terminal equality constraint that is numerically difficult to deal with. Moreover, the online NMPC has more freedom and can deal with cases where the LTO trajectory is not reachable within the prediction horizon.

To highlight that horizon length is a fundamental issue for autonomous racing, note that \cite{Kabzan2019_AMZ} required a 2s look ahead at narrow formula student track and a top speed of 15\unit{m/s}, which is at most a 30m look ahead, and for full-scale cars \cite{Novi2019} required a 400m look ahead.

\subsection{Lap Time Optimization}
\label{section:lto}
The optimal race line is computed by a minimum-time optimal control problem that uses the spatial dynamics \eqref{eq:spatialdomain}.  
To formulate the optimization problem, we use the center line of the pre-mapped race track as a reference path and discretize the continuous space dynamics. We discretize $s$ with a discretization distance of $\Delta s$ and use a spatial forward Euler integrator. Thus, the discrete-space dynamics have the form, 
\begin{equation*}
    \mathbf{{x}}_{k+1} = f_s^d(\mathbf{{x}}_k, \mathbf{u}_k) = \mathbf{{x}}_k + \Delta s f_s^c(\mathbf{{x}}_k,\mathbf{u}_k)\,.
\end{equation*}
To enforce that the resulting optimal trajectory is periodic, a periodicity constraint is inserted, 
\begin{equation*}
 f_s^d(\mathbf{x}_N, \mathbf{u}_N) = \mathbf{x}_{0}\,,
\end{equation*}
where $N$ is the number of discretization steps, and $\Delta s$ is chosen such that $(N+1) \Delta s$ is equal to the arc-length of the center line.

The cost function seeks to minimize the lap time $T = \sum_{k=0}^N \Delta s /\dot{s}_k$. The cost also contains two regularization terms, a slip angle cost, which penalizes the difference between the kinematic and dynamic side slip angle, and a penalization of the rate of change of the physical inputs. The first regularization term is
\begin{equation}
B(\mathbf{x}_k) = q_\beta(\beta_{\rm{dyn},k} - \beta_{\rm{kin},k})^2\,,
\label{eq:slipangle}
\end{equation}
where, $q_\beta$ is a positive weight, $\beta_{\rm{kin},k} = \arctan (\delta_k l_R /(l_F + l_R))$, and $\beta_{\rm{dyn},k} = \arctan (v_{y,k} / v_{x,k})$. The regularizer on the rate of change of the physical inputs is $\mathbf{u}^T R \mathbf{u}$, where $R$ is a diagonal weight matrix. In summary, the overall cost function is,
\begin{equation*}
    j_{\rm{LTO}}(\mathbf{x}_k, \mathbf{u}_k) = \Delta s \frac{1}{\dot{s}_k} + \mathbf{u}^T R \mathbf{u} + B(\mathbf{x}_k) \,.
\end{equation*}

Finally, we use the constraints introduced in Section \ref{sec:constraints}, to formulate the lap time optimization problem,

\begin{equation}
\begin{aligned}
& \underset{\mathbf{{X}},\mathbf{U}}{\text{min}}
    && \sum_{k=0}^{N} j_{\rm{LTO}}(\mathbf{{x}}_k, \mathbf{u}_k) \\
& \; \text{s.t.}
    && \mathbf{{x}}_{k+1} = f_s^d(\mathbf{{x}}_k, \mathbf{u}_k) \\
&&& f_s^d(\mathbf{{x}}_N, \mathbf{u}_N) = \mathbf{{x}}_{0} \\
&&& \mathbf{{x}}_k \in \mathcal{X}_{\text{Track}}, \quad \mathbf{{x}}_k \in \mathcal{X}_{\text{FE}} \\
&&& \mathbf{a}_k \in \mathcal{A}, \quad k = 0,...,N \,,
\label{eq:LTO}
\end{aligned}
\end{equation}
where $\mathbf{{X}} = [\mathbf{{x}}_0,...,\mathbf{{x}}_N]$ and $\mathbf{{U}} = [\mathbf{{u}}_0,...,\mathbf{{u}}_N]$. The problem is formulated in the automatic differentiation package \emph{CppAD} \cite{CppAD} and \emph{Ipopt} \cite{ipopt} is used to solve the resulting nonlinear optimization problem.

\subsection{MPC-Curv}
We use an online NMPC module that we call MPC-Curv, which follows the LTO path. The formulation of the online MPC-Curv problem is very similar to the offline LTO problem, as they share the goal of finishing a lap as fast as possible. However, MPC-Curv uses the time domain model, since it is better suited for online control. Note that our state estimation pipeline, as well as the low-level controllers run in discrete time, additionally, the time domain model is computationally less demanding, as already discussed in \ref{sec:spacevstime}. Instead of minimizing time, MPC-Curv maximizes the progress over the horizon $p = \sum_{t=0}^N \Delta t \; \dot{s}_t$, where $\Delta t$ is the sampling time. 

To further reduce the solve times of the MPC-Curv, we aim to reduce the number of states in the optimization problem. Since only the curvature and the track constraints depend on the progress state $s$, and $s$ can be accurately estimated from the previous MPC solution\footnote{At least for single agent autonomous racing}, MPC-Curv fixes $s$ to the estimated values, and discards it in the dynamics. Note that to get the full benefit from including the $s$ state, one would need to include a parametric curvature function in the model and constraint evaluation of the MPC which would result in additional computational overhead. Thus, we also use  $\mathbf{{x}} = [n;\mu;v_x;v_y;r;\delta;T]$ as the state in the MPC-Curv problem. 

The MPC-Curv cost function combines the progress optimization with the same regularization terms as in the LTO problem \eqref{eq:slipangle}. However, since the goal is to follow the LTO path, two path following terms are added, minimizing the deviation and local heading with respect to the reference path. The MPC-Curv cost function can therefore be formulated as
\begin{equation*}
 j_{\rm{MPC}}(\mathbf{{x}_t}, \mathbf{u}_t) = - \Delta_t\; \dot{s}_t + q_n n_t^2 + q_\mu \mu_t^2 + \mathbf{u}^T R \mathbf{u} + B(\mathbf{{x}}_t)\,,
\end{equation*}
where, $q_n$ and $q_\mu$ are positive path following weights, $R$ a diagonal matrix with positive weights, and $B(\mathbf{x})$ is the slip angle cost (\ref{eq:slipangle}). In addition to the constraints an LTO terminal velocity constraint $v_{x,T} \leq \bar{V}_{x}(s_T)$ is added, where $\bar{V}_{x}$ comes from the LTO trajectory. Thus the MPC-Curv problem is

\begin{equation*}
\begin{aligned}
& \underset{\mathbf{X}, \mathbf{U}}{\text{min}}
    && \sum_{t=0}^{T} j_{\rm{MPC}}(\mathbf{x}_t, \mathbf{u}_t) \\
& \; \text{s.t.}
    && \mathbf{x}_0 = \mathbf{\hat{x}}\\
    &&& \mathbf{x}_{t+1} = f_t^d(\mathbf{x}_t, \mathbf{u}_t) \\
&&& \mathbf{{x}}_t \in \mathcal{X}_{\text{Track}}, \quad \mathbf{{x}}_t \in \mathcal{X}_{\text{FE}} \\
&&& \mathbf{a}_t \in \mathcal{A}, \quad v_{x,T} \leq \bar{V}_{x}(s_T)\\
&&& t = 0,...,T \, ,
\end{aligned}
\end{equation*}
where the subscript $t$ is used to highlight that the problem is formulated in the time domain. Further, $\mathbf{\hat{x}}$ is the current curvilinear state estimate and $T$ is the prediction horizon. The discrete time dynamics $f_t^d({\mathbf{x}, \mathbf{u}})$ is obtained by discretizing the continuous time dynamics with an RK4 integrator. Note that $s_T$ is fixed and is based on the previous MPC-Curv solution, identical to all other occurrences of the progress state $s_t$.

The optimization problem is solved in real-time using ForcesPro, a proprietary interior point method solver optimized for NMPC problems \cite{FORCESPro,FORCESNLP}. 

\section{Results}
\label{sec:results}

The hierarchical controller was tested in simulation and on-track with the full size autonomous racing vehicle, shown in Fig. \ref{fig:pilatus}.
The vehicle is the  AMZ  Driverless 2019 race car, a lightweight single seater race car, that has an all-wheel drive electric powertrain. The  race car  is  equipped  with a complete sensor suite including two LiDARs, three cameras, an optical absolute speed sensor and an INS system, among others. The race car is also equipped with an Intel Xeon E3 that runs the control framework introduced in this paper, but also the mapping, localization and state estimation; see \cite{valls2018design,gosala2019redundant,Kabzan2019_AMZ} for more details. The following results showcase the performance of the controller, both in simulation and during the final racing event at Formula Student Germany (FSG)\footnote{www.formulastudent.de}.

The race track used at the competition is composed of sharp turns, straights and a long sweeping curve. During the competition, the track is first mapped, while a simple pure pursuit controller is used to drive the vehicle. The proposed architecture is then used for all the subsequent runs, based on the map of the track, first the LTO trajectory is computed by solving \eqref{eq:LTO}, and online the MPC-Curv is following the LTO path, as shown in Fig. \ref{fig:diagram}

We first describe the implementation details of the LTO problem in Section \ref{sec:LTO_imp}, in Section \ref{sec:centerVsLTO} we show simulation results that investigate the influence of the reference path and the terminal velocity constraint on the performance of the controller. Next, the experimental results using the proposed controller are presented in a competition setting in Section \ref{sec:exp}. 

\subsection{Computation of the LTO trajectory} \label{sec:LTO_imp}

For the LTO problem, we use $N = 1000$ discretization steps, which corresponds to a discretization distance of $\Delta s = 30.7\unit{cm}$. Finally, we limited the maximum velocity to 17\unit{m/s}, for safety reasons. Given this discretization, the optimization problem \eqref{eq:LTO} is solved in approximately 12s on a i7-7500U processor. The resulting LTO trajectory is shown in Fig. \ref{fig:optimal_race_line}.

\subsection{Following the LTO Trajectory}
\label{sec:centerVsLTO}

\begin{figure}[h]
\centering
\includegraphics[width=0.475\textwidth]{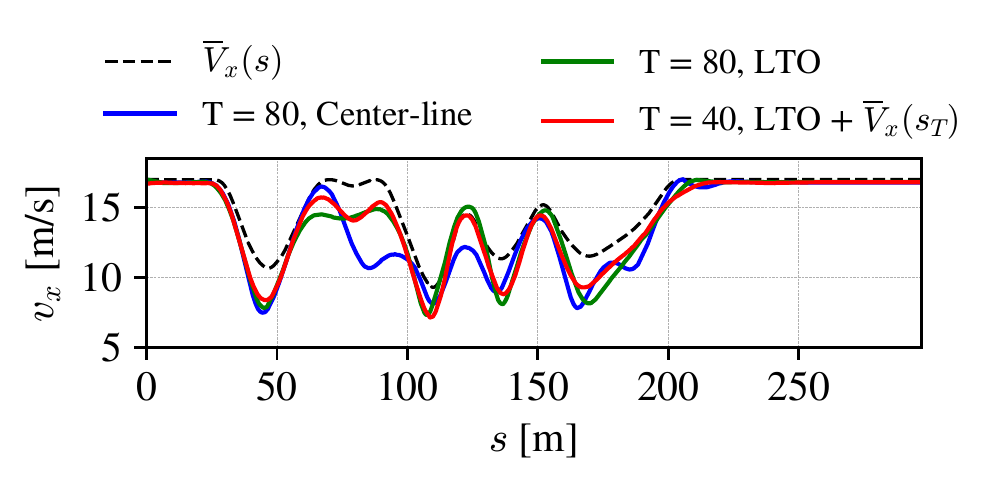}
\caption{Comparison of MPC-Curv speed profiles computed for controller following the center line without terminal constraint on $v_x$ (blue line), controller following the LTO line without terminal constraint (green line), and controller following the LTO line with terminal constraint (red line). The velocity profile computed from LTO is shown for reference (black dashed line).} 
\label{fig:v_x}
\end{figure}

To validate that an ideal line and terminal velocity constraints are beneficial, we tested three different controllers in simulation. First, we tested MPC-Curv \emph{without} terminal velocity constraint, and using the center line as a reference. To compensate for the lack of terminal constraint, we used a longer look-ahead of 2s, which, given our sampling time of $\Delta t = 25$ms results in $T=80$. Note that this setup is comparable to the MPCC used in \cite{Kabzan2019_AMZ}, however MPC-Curv still comes with computational advantages, due to the reduced number of optimization variables. The second controller follows the LTO path, still without terminal constraint. The required horizon in this case is again of 2s. The current approach proposed here, where MPC-Curv follows the LTO path with a terminal constraint, results in a significantly shorter horizon of 1s. The resulting longitudinal velocity for the three MPC-Curv variants, as well as the LTO velocity profile are shown in Fig. \ref{fig:v_x}. The velocity profiles are shown with respect to the arc-length of the center line, and for reference we highlighted the start-finish line in Fig. \ref{fig:optimal_race_line} which is the $s=0$ point. 

We can clearly see that following the LTO path allows the MPC-Curv to drive faster, compared to following the center line. We believe that this is due to the fact that following the LTO path partially overcomes the limitations of a finite look ahead, as it offloads line choice to the higher level LTO problem. This allows reaching higher speeds especially in the complex in-field section of the race track. Fig. \ref{fig:v_x} also clearly shows that using a terminal velocity constraint enables the use of a short 1s look-ahead without sacrificing performance. It comes however with the cost that the brake points are solely dictated by the terminal constraint. The lap times also confirm this discussion, where following the center line results in the slowest lap time of 22.70s, using the LTO path allows for a reduced lap time of 21.67s, with identical lap times for both LTO-based approaches.

Note that from Fig. \ref{fig:v_x} it seems that the online MPC-Curv is not fully achieving the LTO-based speed. However, the difference is due to different tuning, LTO is tuned to give an optimistic velocity profile and MPC-Curv to drive the real car. Future research is needed to get the online and offline phase of our hierarchical controller closer together. 

\subsection{Experimental Validation}
\label{sec:exp}
\begin{figure*}[t]
    \centering
    \includegraphics[width=0.98\textwidth]{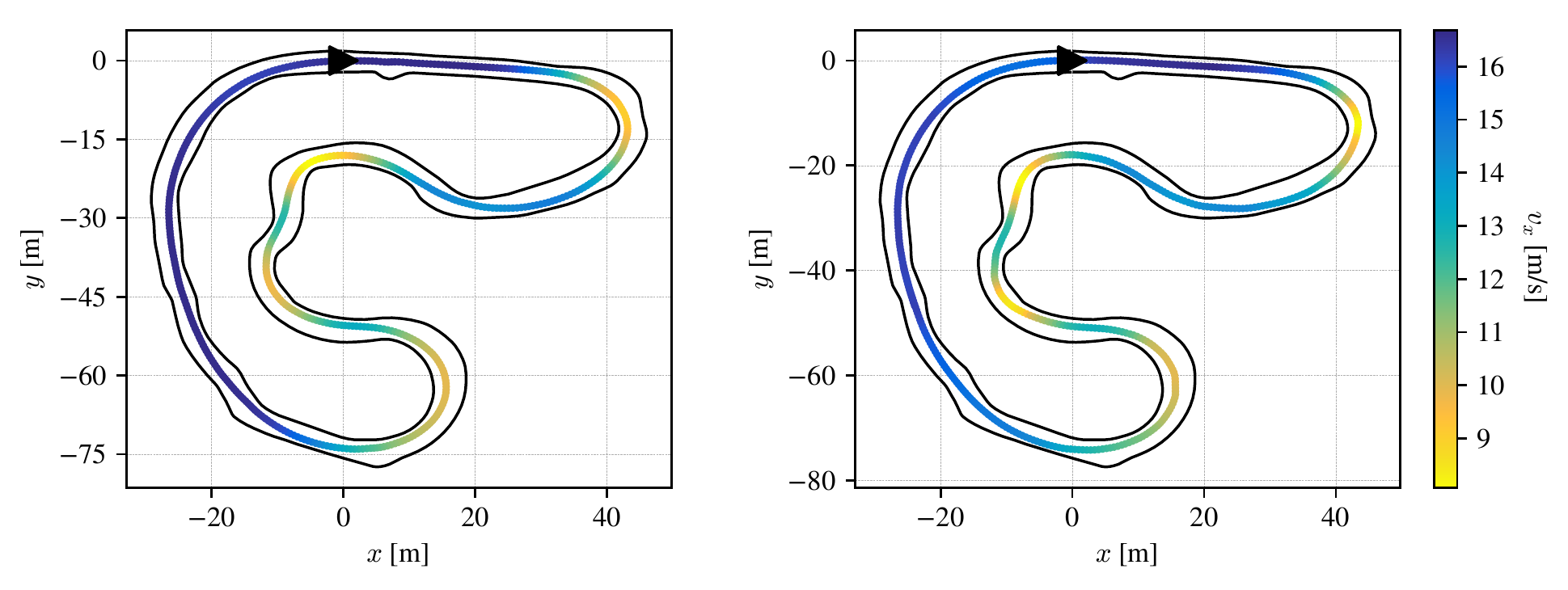}
    \caption{The racing trajectory calculated by the lap time optimization (left panel) and the real-time trajectory of the vehicle (right panel). The information about $v_x$ is given by the color of the trajectory.}
    \label{fig:optimal_race_line}
\end{figure*}

The experimental validation of the hierarchical controller was done during the Formula Student Germany Driverless 2019 competing. The controller at the competition used a look-ahead of 1s with a sampling time of $\Delta t = 25$ms, and a horizon length of $T=40$, identical to the third controller discussed in Section \ref{sec:centerVsLTO}. The LTO path and velocity profile used as a reference path and terminal velocity constraint are shown in Fig. \ref{fig:optimal_race_line}. 

One advantage of the proposed control structure is that compared to the current state of the art method \cite{Kabzan2019_AMZ}, the optimization variables per time step are reduced. This allows us to achieve computation times of 12.81ms in average, with only 3\% of the time instances over the 25ms limit. Compared to \cite{Kabzan2019_AMZ} the decrease in computation time is approximately \emph{three}-fold, which is a drastic enhancement, since the same solver and prediction horizon are used. Note that a different processor is used, but even with identical processors, we noticed at least a two-fold decrease in computation times.

Fig. \ref{fig:optimal_race_line} shows that the LTO path (left) and the experimental results (right) are generally very similar. Therefore, MPC-Curv is able to follow the LTO path accurately. The main difference is that the experimental velocity slightly lags behind the LTO velocity. Since this is not the case in the simulation results, the cause is most likely a mismatch between the model used in the MPC and the dynamics of the real race car. Note that this mismatch also caused the car to slightly miss the apex in the top right curve.

When comparing the time to finish the 10 lap race at the competition, our proposed hierarchical controller achieved an average lap time of 22.63s. This is 20\% faster than the average lap time of the 2018 competition winner \cite{Kabzan2019_AMZ}, which achieved a lap time of 28.59s. Note that the track at the two competitions was very similar but not identical and slightly more complex in 2018. Our simulations suggests that the difference between the tracks is in the order of 1.5s. The average lap time was also 8\% faster, than the second fastest car at the 2019 competition which achieved an average lap time of 24.49s. We would like to note that 8\% in terms of lap time is still a large margin. For a video of the 10 lap race see \url{https://youtu.be/gcnngFyWnFQ?t=13079}.

\begin{figure}[h]
    \centering
    \includegraphics[width=0.45\textwidth]{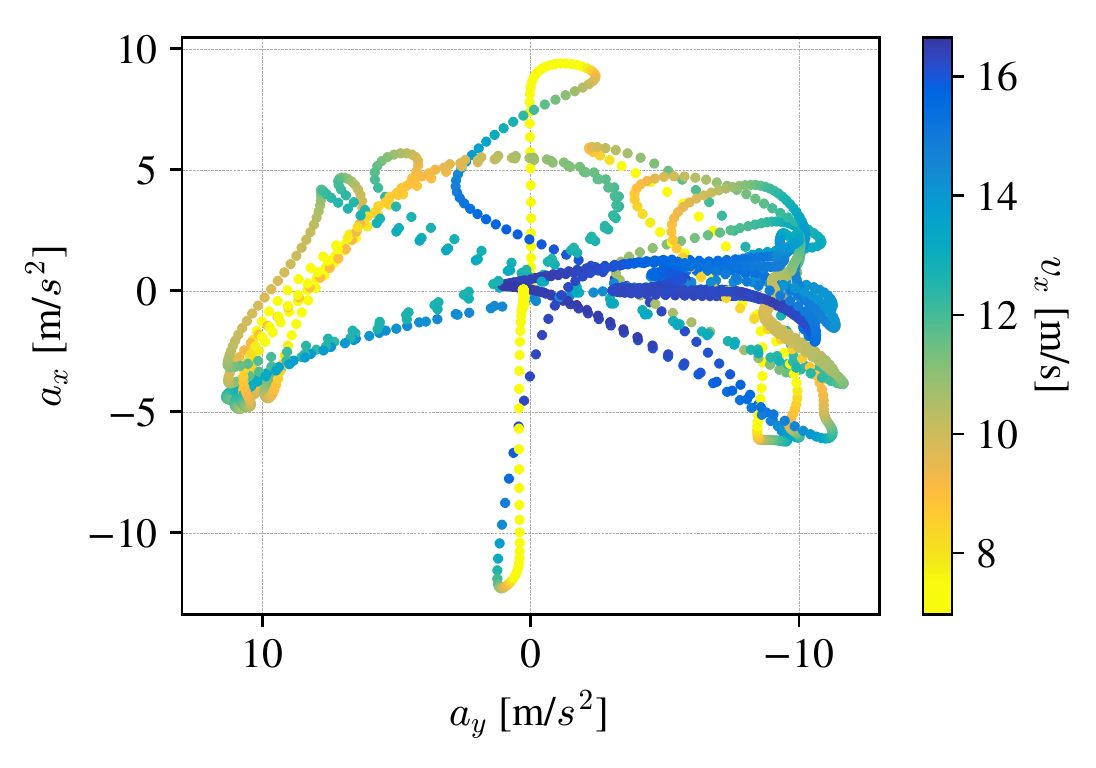}
    \caption{Longitudinal and lateral accelerations during Formula Student Germany.}
    \label{fig:gg_plot}
\end{figure}

Finally, Fig. \ref{fig:gg_plot} shows the combined acceleration in lateral and longitudinal direction (GG-diagram) of the run at Formula Student Germany competition, where it can be seen that the car achieves lateral accelerations of up to $12\unit{m/s^2}$. Compared to the performance of novice race car drivers this can be considered as impressive, however expert drivers would be able to achieve lateral acceleration of over $18\unit{m/s^2}$. %see Appendix \ref{sec:appendix}.

\section{Conclusion}
\label{sec:conclusion}
A novel control approach, consisting of a hierarchical MPC using a bicycle model in a curvilinear coordinate system was presented. The high-level controller computes a trajectory that minimizes the lap time, and the low-level nonlinear model predictive path following controller tracks the computed trajectory online. The two levels are further coupled through a terminal constraint, computed in the high-level controller, and used in the online optimization by the low-level controller which ensures that while short prediction horizons can be used online, safety is maintained at all times. The framework was tested on a full-size Formula Student race car, demonstrating significant performance improvements in computation time and lap time reduction, as compared to the current state of the art results achieved on similar platform and track. Further research will address improvements in the coupling of the higher and the lower level controllers, and investigating model learning \cite{Kabzan2019_learning} which we believe will bring us close to human expert performance. 

\section*{Acknowledgement}
We would like to thank the entire AMZ Driverless team, this work would not have been possible without the effort of every single member, we are glad for having the opportunity to work among such amazing people. 
\balance
\bibliography{bib}
\bibliographystyle{IEEEtran}

\end{document}